\pdfoutput=1
\documentclass[11pt]{article}
\usepackage[final]{acl}
\usepackage{booktabs}
\usepackage{times}
\usepackage{latexsym}
\usepackage{amssymb}
\usepackage{subcaption}
\usepackage{graphicx}
\usepackage{tabularx}
\usepackage{multirow}
\usepackage{caption}
\usepackage[T1]{fontenc}
\usepackage{amsmath}
\usepackage[utf8]{inputenc}
\usepackage{enumitem}
\usepackage{microtype}
\usepackage{rotating}
\usepackage{inconsolata}

\usepackage{graphicx}
\usepackage{xcolor}
\definecolor{red}{RGB}{216,30,6}
\definecolor{blue}{RGB}{18,150,219}

\author{
    Yue Li\textsuperscript{\rm 1}\thanks{Equal Contribution.}, Xin Yi\textsuperscript{\rm 1}\footnotemark[1], Dongsheng Shi\textsuperscript{\rm 1}, Gerard de Melo\textsuperscript{\rm 2} \\ 
    \textbf{Xiaoling Wang$^{1}$, Linlin Wang$^{1}$\thanks{Corresponding Author.}} \\
    $^{1}$East China Normal University\\
    $^{2}$Hasso Plattner Institute/University of Potsdam\\
    \texttt{\{yue\_li,xinyi,dongsheng\}@stu.ecnu.edu.cn, gdm@demelo.org,} \\ \texttt{ \{xlwang,llwang\}@cs.ecnu.edu.cn} \\
}

\title{Hierarchical Safety Realignment: Lightweight Restoration of Safety\\in Pruned Large Vision-Language Models}
\begin{document}
\maketitle
\begin{abstract}
With the growing size of Large Vision-Language Models (LVLMs), network pruning techniques designed to compress these models for deployment in resource-constrained environments have attracted significant attention. However, we observe that pruning frequently results in a degradation in safety performance. To address this issue, we propose a novel and lightweight approach, named \textbf{H}ierarchical \textbf{S}afety \textbf{R}ealignment (\textbf{HSR}). HSR operates by first quantifying the contribution of each attention head to safety, identifying the most critical ones, and then selectively restoring neurons directly within these attention heads that play a pivotal role in maintaining safety. This process hierarchically realigns the safety of pruned LVLMs, progressing from the attention head level to the neuron level. We validate HSR across various models and pruning strategies, consistently achieving notable improvements in safety performance. To the best of our knowledge, this is the first work explicitly focused on restoring safety in LVLMs post-pruning. The code will be available at \url{https://github.com/TheShineyue/HSR}.
\end{abstract}

\section{Introduction}

Large Language Models (LLMs) benefit from their massive parameter count and advanced architectures, achieving outstanding results on diverse benchmarks. Building on this success, efforts to extend LLMs into multimodal domains have made remarkable progress as well.  Most current Large Vision-Language Models (LVLMs), typically composed of visual encoders, adapters, and LLM backbones \citep{liu2024survey}, 
have a large parameter scale and leverage image-text datasets to achieve effective multimodal alignment. To enable model deployment and application under resource-constrained environments, pruning methods \citep{sun2024a, frantar2023sparsegpt,lee2018snip} compute importance scores of neurons to eliminate those deemed less important, thereby reducing the size of the model while retaining the utility to a certain 
extent. Such methods have seen broad adoption to boost model efficiency.
%This line of work has gained widespread adoption for enhancing model efficiency.

\begin{figure}[t]
  \includegraphics[width=\columnwidth]{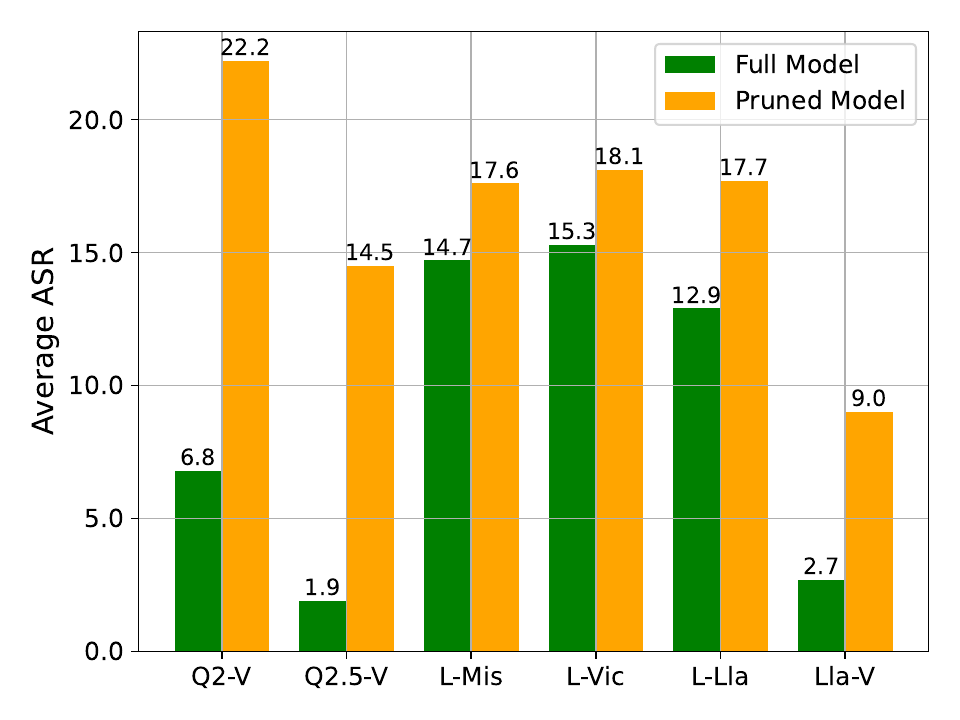}
  \caption{The Average ASR (lower values indicate higher safety) of full baselines versus pruned models (50\% sparsity) across safety evaluation benchmarks. The abbreviations on the x-axis, listed from left to right, correspond to the following models: Qwen2-VL, Qwen2.5-VL, LLaVA-NeXT-Mistral, LLaVA-NeXT-Vicuna, LLaVA-NeXT-Llama3, and Llama3.2-Vision.}
  \label{fig:prune_full}
\end{figure}

Safety is commonly defined as preventing models from following malicious instructions and generating toxic content \citep{bianchisafety}. Recent work \citep{zhou2025on} has revealed that certain safety heads within the attention mechanism are crucial for feature integration in safety-related  tasks. Additionally, neural-level research \citep{wei2024assessing} has found that certain regions within the model are critical for safety guardrails, which are separate from the utility-related regions and exhibit sparsity. A natural concern arises from the fact that, because these regions contribute minimally to utility, they are prone to being removed by pruning technologies that prioritize utility importance as a pruning metric. This removal could result in a decline in the safety of the pruned model.
To verify whether this problem exists, we used the Wanda pruning method \citep{sun2024a} to prune six mainstream LVLMs and compared their safety changes before and after pruning. The experimental results shown in Figure~\ref{fig:prune_full} reveal that all LVLMs exhibited varying degrees of safety degradation, with the worst-performing model showing a 15.4\% safety drop and the best-performing case exhibiting a 2.8\% decline. Despite the serious safety risks of pruning technologies, research on model safety restoration after pruning remains scarce.

To address this problem, we propose a novel \textbf{H}ierarchical \textbf{S}afety \textbf{R}ealignment (\textbf{HSR}) approach, designed to restore the safety performance degraded by pruning, without significantly increasing the pruned model's parameter size. Our HSR  apporach hierarchically realigns the safety of the pruned model from attention head level to  neuron level. HSR operates in two main steps: First, we evaluate each attention head's contribution to model safety and identify the safety-critical heads with the greatest impact. Subsequently, for these key heads, we pinpoint and restore safety-critical neurons that were pruned, effectively realigning the safety of our model.

We have validated the effectiveness of our approach across various models and pruning techniques. Our proposed HSR approach successfully realigns the safety of pruned models, restoring over 27\% of the lost safety in many cases and more than 14\% even in the worst-case scenarios, all with lightweight modifications. Furthermore, with extensive analysis and ablation experiments, we have uncovered several key insights into model safety. These include the finding that a small subset of neurons plays a disproportionately significant role in ensuring safety, and the observation that certain neurons exist which negatively impact safety. 

In summary, our contributions are the follows: 

\begin{itemize}
    \item 
    We propose a novel method named Hierarchical Safety Realignment (HSR) to realign the safety of pruned LVLMs, yielding substantial safety improvements with lightweight modifications. To our knowledge, HSR is the first method specifically designed to address the safety realignment of pruned LVLMs.

    %We validate the proposed method across various LVLMs with different pruning techniques. Extensive experiments demonstrate the superiority of our approach, showing consistent performance improvements with minimal neuron restoration.

    \item  Our findings reveal that a small subset of neurons plays a disproportionately large role in ensuring safety, while certain neurons negatively impact safety. By selectively restoring these safety-critical neurons, we can achieve significant safety recovery in pruned models.

    \item We validate the proposed method on various LVLMs using different pruning techniques. Extensive experiments demonstrate the superiority of our approach, consistently improving performance with minimal neuron restoration.
    
\end{itemize}

\section{Method}
\begin{figure*}[t]
  \includegraphics[width=\textwidth]{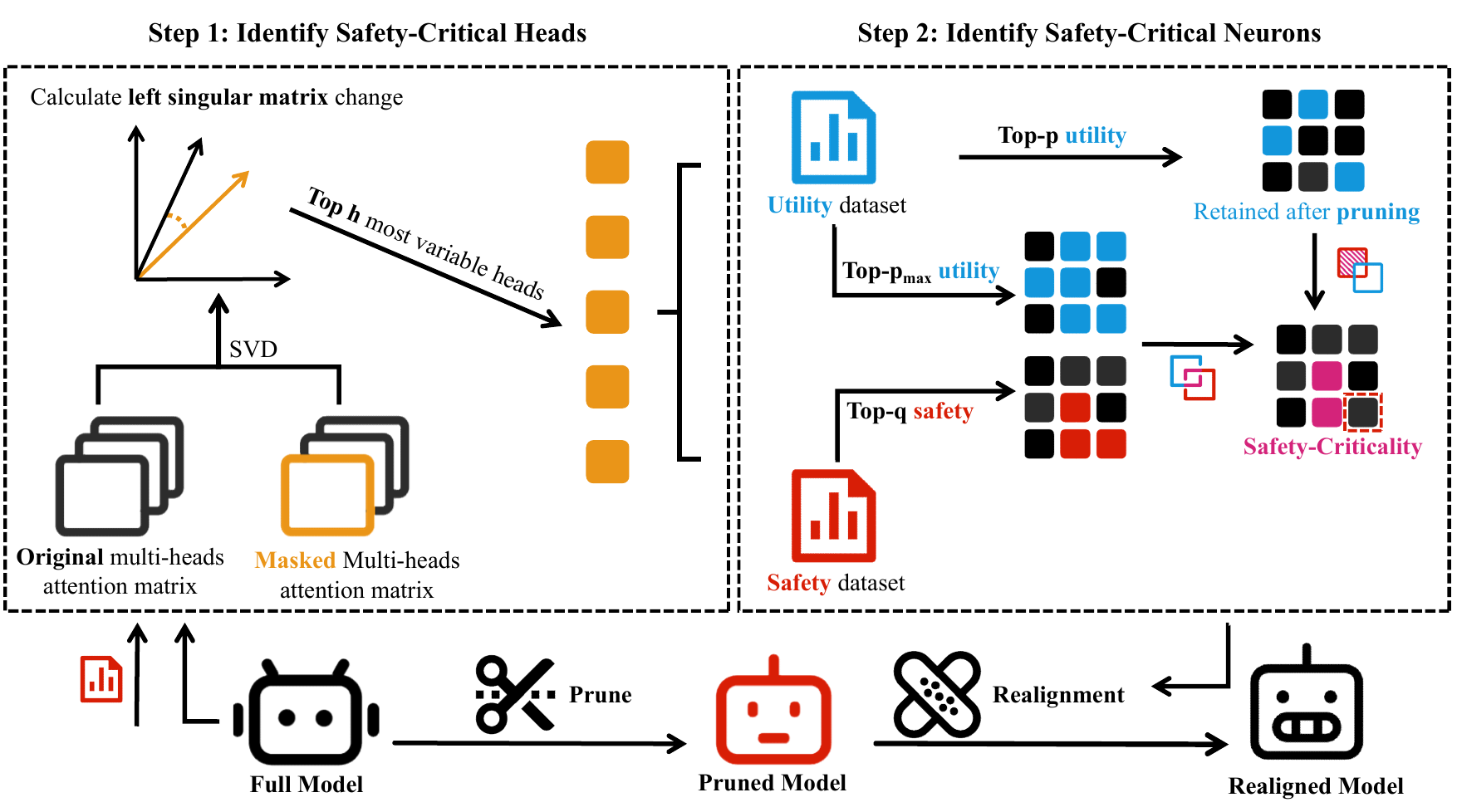}
  \caption{HSR hierarchically achieves safety realignment of the pruned model in two steps: The first step identifies the top-h most important attention heads for safety, while the second one identifies and restore the safety-critical neurons on these heads. The safety dataset, comprising malicious instructions and appropriate rejection responses, is marked in \textcolor{red}{red}, and the utility dataset, which excludes malicious instructions, is marked in \textcolor{blue}{blue}.
  }
  \label{fig:Method}
\end{figure*}
\textbf{Overview}\quad In this section, we present our Hierarchical Safety Realignment method in detail and illustrate the core process in Figure~\ref{fig:Method}. Our approach achieves safety realignment of the pruned model by hierarchically identifying and restoring safety-critical neurons, starting at the attention head level and progressing to the neuron level:
At the \textbf{attention head} level (Section~\ref{subsec:heads}), each attention head is individually masked to measure changes in the model’s output for malicious instructions compared to the original. The attention heads causing the most significant changes, termed as the \textbf{safety-critical} heads, are selected for further analysis. 
At the \textbf{neuron} level (Section~\ref{subsec:neuron}), we compute two importance scores for neurons: the safety importance score based on a safety dataset and the utility importance score derived from a utility dataset (details provided in the caption of Figure~\ref{fig:Method}). Pruned neurons exhibiting high safety importance along with sufficient utility importance are identified as \textbf{safety-critical} neurons and subsequently restored. 

\subsection{Identifying Safety-Critical Heads}

\label{subsec:heads}
To preserve model sparsity, we selectively restore neurons only in those attention heads that exhibit the highest safety-critical importance. \citeposs{zhou2025on} work proposed a new metric tailored for multi-head attention, namely the \textbf{S}afety \textbf{h}ead \textbf{i}m\textbf{p}ortance \textbf{s}core (\textbf{Ships}), to evaluate the contribution of each head to the model safety. 
Specifically, for a specific harmful data $q_{\mathcal{H}}$, the probability distribution of the original model $\theta_{\mathcal{O}}$ is denoted by $p(q_{\mathcal{H}}; \theta_{\mathcal{O}})$. For the $i$-th attention head $h_i^l$ in the $l$-th layer, its contribution to the safety of the model is eliminated by multiplying its Query, Key, and Value matrices by a very small coefficient $\epsilon$. The probability distribution of the model after such ablation is denoted by $p(q_{\mathcal{H}}; \theta_{\mathcal{O}} \setminus \theta_{h_i^l})$. The $\text{Ships}(q_{\mathcal{H}},\theta_{h_i^l})$ are calculated as the KL divergence \citep{kullback1951information} of $p(q_{\mathcal{H}}; \theta_{\mathcal{O}})$ and $p(q_{\mathcal{H}}; \theta_{\mathcal{O}} \setminus \theta_{h_i^l})$ as follows:%shown in Eq. 1The KL divergence \citep{kullback1951information} between $p(q_{\mathcal{H}}; \theta_{\mathcal{O}})$ and $p(q_{\mathcal{H}}; \theta_{\mathcal{O}} \setminus \theta_{h_i^l})$, operationalized through Equation~\ref{eq:ships}
\begin{equation}
  \label{eq:ships}
  % \text{Ships}(q_{\mathcal{H}},\theta_{h_i^l}) = 
     \mathbb{D}_{KL} \left( p(q_{\mathcal{H}}; \theta_{\mathcal{O}}) \parallel p(q_{\mathcal{H}}; \theta_{\mathcal{O}} \setminus \theta_{h_i^l}) \right)
\end{equation}
%named $\text{Ships}(q_{\mathcal{H}},\theta_{h_i^l})$. 
It quantifies the impact of ablating head $h_i^l$ for $q_{\mathcal{H}}$, which is the safety contribution of $h_i^l$.
%Specifically, for $i$-th attention head $h_i$, the contribution of $h_i$ is ablated by multiplying the matrix corresponding to $\mathbf{W}_q$, $\mathbf{W}_k$ and $\mathbf{W}_v$ by a very small coefficient $\epsilon$. % 具体而言，对于一个有害数据a,原始模型sita第l层的概率分布是xxx。对第l层的第i个注意力头hi，通过将其Query, Key, and Value matrices乘上一个极小系数e消融其对模型安全性的贡献，此时消融后模型的概率分布为。通过函数1计算原有概率分布x和消融后的概率分布xx的变化给出其对安全性的贡献, 即Ship。The change between the $p(q_{\mathcal{H}}; \theta_{\mathcal{O}})$ and $p(q_{\mathcal{H}}; \theta_{\mathcal{O}} \setminus \theta_{h_i^l})$ is calculated by Equation~\ref{eq:ships} to give the contribution of $h_i^l$ to safety.
Given that most contemporary LVLMs employ Group Query Attention \citep{ainslie2023gqa} to reduce computational overhead, we derive generalized masking equations. For the query and key matrices $\mathbf{W}_q$ and $\mathbf{W}_k$, the modified head $h_i^m$ calculation becomes:
% Considering that most of the current mainstream LVLMs use the Group Query Attention \citep{ainslie2023gqa} mechanism to save time overhead, we give more general mask equations. For the $\mathbf{W}_q$ and $\mathbf{W}_k$ matrices, we calculate the modified $h_i^m$ as follows:
\begin{equation}
  \label{eq:gqa_qk}
     h_i^m = \text{Softmax}\left(\frac{\epsilon \mathbf{W}_q^i {\mathbf{W}_k^{i/g}}^T}{\sqrt{d_k/n}}\right) \mathbf{W}_v^{i/g},
\end{equation}
% additionally for the $\mathbf{W}_v$ matrix, we calculate as follows:
whereas for the value matrix $\mathbf{W}_v$, the calculation is adjusted as:
\begin{equation}
  \label{eq:gqa_v}
     h_i^m = \text{Softmax}\left(\frac{\mathbf{W}_q^i {\mathbf{W}_k^{i/g}}^T}{\sqrt{d_k/n}}\right) \epsilon \mathbf{W}_v^{i/g},
\end{equation}
Here, $n$ denotes the number of attention heads per layer, and $g$ denotes the query amount of each group calculated as $g = n/n_{kv}$ where $n_{kv}$ indicates the number of key-value head pairs.

For a given dataset $D$, we aggregate network activations into matrix $\mathbf{X}$, and perform singular value decomposition (SVD): $\text{SVD}(\mathbf{X}) = \mathbf{U}\Sigma \mathbf{V}^\top$, where $\mathbf{U}$ represents the key features in the dataset space. Through this decomposition, we derive two critical matrices: $\mathbf{U}_\theta$ (left feature matrix from the original model) and $\mathbf{U}_A$ (left feature matrix from the ablated model). The safety representation divergence is quantified using the $\text{Ships}(D, h_i^l)$ metric:  

\begin{equation}
  \label{eq:Sahara}
     \text{Ships}(D, h_i^l)= \sum_{r=1}^{r_{\text{max}}} \cos^{-1} \left( \sigma_r(U_\theta^{(r)}, U_A^{(r)}) \right)
\end{equation} where $\sigma_r$ denotes the $r$-th singular value. A larger main angle indicates that the safety representation has changed significantly, which represents the safety importance at the dataset level. We subsequently identify the top-h attention heads with maximal safety contributions, designated as \textbf{safety-critical} heads, for neuron-level attribution analysis.

\subsection{Identifying Safety-critical Neurons}
\label{subsec:neuron}

\subsubsection{Quantifying Neuron Importance}
\label{subsubsec:wanda_sparsegpt_snip}
We proceed to identify pruned neurons that remain critical for safety considerations.
Given a calibration dataset, the pruning method calculates importance scores of weights to attribute their impact on the model's relative performance. When provided with a safety dataset or a utility dataset, the method quantifies the safety importance scores and utility importance scores of the weights, respectively. We provide three variants that use different approaches to quantify neuron importance as follows:

% 1. We leverage xxx score to ...
\begin{itemize}
\item For a given calibration dataset, we use Wanda Score \citep{sun2024a} to calculate the importance score of a weight using the absolute value of its weight matrix and the $\ell_2$ norm of the input activations. Subsequently, we follow \citealp{wei2024assessing} to mask the rest of the calibration dataset and focusing only on the response activation, and tore all activations for layer W into $\mathbf{X}_{\text{in}}$ of shape $(n , C_{\text{in}})$ and calculate the importance score $\mathbf{I}$ as:

\begin{equation}
  \label{eq:wanda_plus}
    \mathbf{I} = |\mathbf{W}| \odot \left( \mathbf{1} \cdot \| \mathbf{X}_{\text{in}} \|_2^\top \right),
\end{equation}
where $ |\mathbf{W}| $ is a weight matrix  of 
a linear layer  of shape $(C_{\text{out}}, C_{\text{in}})$. $\mathbf{1}$ denotes an all-one vector of shape $(C_{\text{out}}, 1)$. We compute the row-wise $\ell_2$ norms of $ \mathbf{X}_{\text{in}}$, and then transpose them to obtain a matrix of shape $(1,C_{\text{in}})$.
\item  We use SparseGPT Score \citep{frantar2023sparsegpt} to obtain the importance $\mathbf{I}$ as Eq. \ref{eq:sparseGPT_plus}, where $\mathbf{X}_{\text{in}}$ contains only response activations:

\begin{equation}
  \label{eq:sparseGPT_plus}
    \mathbf{I} = \left[ \frac{|\mathbf{W}|^2}{\operatorname{diag}\left( (\mathbf{X}_{\text{in}}^\top \mathbf{X}_{\text{in}} + \lambda \mathbf{I})^{-1} \right)} \right]
\end{equation}
Here, $\mathbf{X}_{\text{in}}^T \mathbf{X}_{\text{in}} + \lambda \mathbf{I}$ in the denominator is the Hessian $\mathbf{H}$ for the layer-wise reconstruction problem and $\lambda$ is the Hessian dampening factor to avoid the collapse of inverse computation. Once $\mathbf{I}$ is calculated, SparseGPT updates the weights by masking less important portions based on the desired sparsity.

\item We finally introduce the third method, which is based on the SNIP Score \citep{lee2019snip}. For a data instance $x = (x_{\text{prompt}}, x_{\text{response}})$, we define the corresponding  
 loss function as the conditional negative log-likelihood $\mathcal{L}(x) = -\log p(x_{\text{response}} \mid x_{\text{prompt}})$. For a weight matrix $\mathbf{W}$, we use SNIP Score to calculate 
its importance score $\mathbf{I}$ as follows:
\begin{equation}
  \label{eq:snip}
    \mathbf{I}(\mathbf{W}_{ij}, x) = |\mathbf{W}_{ij} \cdot \nabla_{\mathbf{W}_{ij}} \mathcal{L}(x)|,
\end{equation}
This equation is the first-order Taylor approximation to the change of the loss when the weight entry $\mathbf{W}_{ij}$ is set to zero. 

Following the experimental setup described by \citet{wei2024assessing}, for a given calibration dataset $D$, we use 
\begin{equation}
  \label{eq:snip_plus_d}
    \mathbf{I} = \mathbb{E}_{x \sim D} \mathbf{I}(\mathbf{W}, x) = \mathbb{E}_{x \sim D} |\mathbf{W} \odot \nabla_\mathbf{W} \mathcal{L}(x)|.
\end{equation}

\end{itemize}

\subsubsection{Safety-Critical Neuron Restoration}
% \textbf{Identification}\quad 
Considering two different calibration datasets, a safety dataset $D^s$ and a utility dataset $D^u$. $D^s$ comprises instructions and images that contain harmful information, along with responses that correctly refuse such information. In contrast, $D^u$ consists of safe instructions and images paired with reasonable responses. Therefore, the safety importance score $\mathbf{I}^s$ and utility importance score $\mathbf{I}^u$ can be calculated respectively using the Wanda, SparseGPT or SNIP method in Section~\ref{subsubsec:wanda_sparsegpt_snip}. % SNIP

We select those weights with larger importance scores and consider them weights that contribute more to safety or utility. Specifically, given hyper-parameters q and p for safety and utility, respectively, we use Equations~\ref{eq:Ss} and~\ref{eq:Su} to obtain the safety importance set $S^{s}$ and utility importance set $S^u$ of the $i$-th layer.
\begin{equation}
  \label{eq:Ss}
    % I^u(p) = \{(i, j) \mid S_{i,j}^u \in \text{Top}_p(S_i^u)\}
    S^{s}(\text{q}) = \{(i, j) \mid \mathbf{I}^u_{i,j} \text{ is the top } \text{q}\% \text{ of } \mathbf{I}^s_i\}
\end{equation}
\begin{equation}
  \label{eq:Su}
    % I^u(p) = \{(i, j) \mid S_{i,j}^u \in \text{Top}_p(S_i^u)\}
    S^{u}(\text{p}) = \{(i, j) \mid \mathbf{I}^u_{i,j} \text{ is the top } \text{p}\% \text{ of } \mathbf{I}^u_i\}
\end{equation}
For the pruning process, the weights outside the utility important set (here p = 1 - sparsity ratio) will be set to 0 according to the set sparsity, thus obtaining a sparse neural network. 

Among the pruned neurons (not in $S^{u}(\text{p})$), we seek those safety-critical neurons that have high safety and still have certain utility, and will not cause excessive loss of model utility in the subsequent realignment process. Therefore, we introduce the hyper-parameter $\text{p}_{\max}$ ($\text{p}_{\max}$ is greater than p) and obtain the safety-critical neurons $S(\text{p},\text{q},\text{p}_{\max})$ as follows:
\begin{equation}
  \label{eq:safe-c}
    S(\text{p},\text{q},\text{p}_{\max}) = ( S^s(\text{q}) \cap S^u(\text{p}_{\max}) ) -  S^u(\text{p}).
\end{equation}
We restore these pruned safety-critical neurons on the pruned model to realign the model’s safety.

\section{Experimental Setup} 
\paragraph{Dataset} During the realignment phase, we employ two distinct data subsets: (1) Safe-Safe pairs (safe images with corresponding safe instructions) 
\begin{table*}
  \centering
  \small
  \newcolumntype{C}{>{\centering\arraybackslash}X}
  \begin{tabularx}{\textwidth}{lCCCCCCCc}
    \toprule
    \multirow{2}{*}{\textbf{Method}}  & \multicolumn{4}{c}{\textbf{Safety $\downarrow$}} & \multicolumn{3}{c}{\textbf{Utility $\uparrow$}} & \multirow{2}{*}{\textbf{Restoration}} \\
    \cmidrule(lr){2-5} \cmidrule(lr){6-8}
    & \textbf{SafeBench} & \textbf{C$h^{3}$Ef} & \textbf{AVG} & \textbf{RSR} & \textbf{MMbench} & \textbf{DocVQA} & \textbf{AVG} \\
    \midrule
    Full Model & 1.40 &	2.35 & 1.88 & - & 87.02 & 94.51 & 90.76  & - \\
    \midrule
    SNIP & 4.60  & 8.12  & 6.36 & - & 84.55 &\textbf{92.93}&88.74& - \\
    \hspace{1em} w/ HSR(Ours) & \textbf{3.00}&\textbf{5.34}&\textbf{4.17}& 48.88\% & \textbf{84.62}&92.90 &\textbf{88.76}  & 0.150‱ \\
    % \midrule
    Wanda & 11.20  &17.74 &14.47 & - & \textbf{85.15}&91.97&88.56& - \\
    \hspace{1em} w/ HSR(Ours) & \textbf{9.00} &\textbf{13.03} &\textbf{11.02}  & 27.40\% & 85.01 &	\textbf{92.13} &\textbf{88.57} & 0.020‱ \\
    % \midrule
    SparseGPT & 3.00 &3.21  &3.10   & - & \textbf{83.88}&	\textbf{90.64} &\textbf{87.26} & - \\
    \hspace{1em} w/ HSR(Ours) & \textbf{2.80}&\textbf{2.56} &\textbf{2.68}  & 34.43\% & \textbf{83.88} &90.63&87.25  & 0.133‱ \\
    \bottomrule
  \end{tabularx}
  \caption{The safety and utility values of Qwen2.5-VL under different pruning methods are shown. Here the Restoration Indicates the ratio of the additional restored parameters to the total parameters that need to be pruned, in ten thousandths. The better value for each group is shown in \textbf{bold}.} 
  \label{tab:main}
\end{table*}
as the utility dataset, and (2) Unsafe-Unsafe pairs (unsafe images with matching unsafe instructions), which constitute the safety dataset. Both subsets are derived from the VLGuard \citep{zong2024safety} training dataset.

During the evaluation phase, we employ the following benchmarks:
(1) For utility assessment, MMbench \citep{liu2025mmbench} and DocVQA \citep{mathew2021docvqa} are utilized;
(2) For safety evaluation, safebench-mini \citep{ying2024safebench} and the harmful subset of $\text{C}h^{3}\text{Ef}$ \citep{shi2024assessment} are adopted.
To ensure fair comparison and reproducibility, all evaluations are conducted under strict zero-shot settings with greedy decoding strategies.

\paragraph{Models for Pruning} Our experiments involved six mainstream LVLMs, including three variants of LLaVA-NeXT\footnote{\url{https://llava-vl.github.io/blog/2024-05-10-llava-next-stronger-llms/}}, built on different language models: Vicuna, Mistral, and Llama3. Additionally, we evaluated Qwen2.5-VL\footnote{\url{https://huggingface.co/Qwen/Qwen2.5-VL-7B-Instruct}}, Qwen2-VL \citep{Qwen2VL}, and Llama-3.2-Vision\footnote{\url{https://huggingface.co/meta-llama/Llama-3.2-11B-Vision-Instruct}}. All models have parameter counts ranging from 7B to 11B.

\paragraph{Evaluation Metrics} Three evaluation metrics we use are Attack Success Rate (ASR), Average Normalized Levenshtein Similarity (ANLS) and Accuracy (Acc). ASR is used to evaluate the safety of the model. The smaller the ASR, the better the safety. We use Llama-Guard-3-Vision\footnote{\url{https://huggingface.co/meta-llama/Llama-Guard-3-11B-Vision}} to determine whether the response is safe. We use ANLS and Acc as the evaluation indicators of DocVQA and MMbench respectively. Following \citeposs{mathew2021docvqa} proposal, ANLS can ensure that minor answer mismatches stemming from OCR errors are not severely penalized. Additionally, in order to more intuitively and fairly reflect the performance improvement of our method, we  provide the \textbf{R}atio of model \textbf{S}afety \textbf{R}ealignment (\textbf{RSR}) as follows: 
\begin{equation}
  \label{eq:safety_re}
\text{RSR} = \frac{
        \text{ASR}_{\text{Pruned}} - \text{ASR}_{\text{Pruned w/ HSR}}
    }
    {
        \text{ASR}_{\text{Pruned}} - \text{ASR}_{\text{Full}}
    }
    %\cdot 100\%
\end{equation}% 为了更直观公平的反应我们
to quantify the ratio of restored safety to lost safety. More relevant details are reported in  Appendix ~\ref{sec:Experimental Details}.

\section{Experimental Results}

\paragraph{Comparison of different pruning methods} We present the performance of Qwen2.5-VL at 50\% sparsity using various pruning methods realigned with HSR, as shown in Table~\ref{tab:main}. HSR effectively realigns the safety of the pruned model and only requires restoring a minimal number of safety-critical neurons. Specifically for Qwen2.5-VL, the average value of ASR decreases by 2.19\%, 3.45\%, and 0.42\%, respectively, for SNIP, Wanda, and SparseGPT. Judging from the ratio of restored safety to lost safety, it is generally possible to recover over 27\% of the safety capacity. In addition, there is no significant loss in utility. In fact, we even find a slight improvement in multiple cases and in the average value, which may stem from the fact that the restored neurons also have a certain utility contribution for Qwen2.5-VL.

\begin{table*}
  \centering
  \small
  \newcolumntype{C}{>{\centering\arraybackslash}X}
  \begin{tabularx}{\textwidth}{lCCCCCCCc}
    \toprule
    \multirow{2}{*}{\textbf{Method}}  & \multicolumn{4}{c}{\textbf{Safety $\downarrow$}} & \multicolumn{3}{c}{\textbf{Utility $\uparrow$}} & \multirow{2}{*}{\textbf{Restoration}} \\
    \cmidrule(lr){2-5} \cmidrule(lr){6-8}
    & \textbf{SafeBench} & \textbf{C}$h^{3}$\textbf{Ef} & \textbf{AVG} & \textbf{RSR} & \textbf{MMbench} & \textbf{DocVQA} & \textbf{AVG} \\
    \midrule
    Qwen2-VL & 5.00 & 8.55 & 6.77 & - & 82.70 & 89.14 & 85.92 & - \\
    \quad Wanda & 21.40 & 23.08 & 22.24 & - & 75.93 & 76.27 & 76.10 & - \\
    \quad w/ HSR (Ours) & \textbf{15.40} & \textbf{18.16} & \textbf{16.78} & 35.29\% & \textbf{76.69} & \textbf{77.93} & \textbf{77.31} & 0.016‱ \\
    LLaVA-NeXT-Mistral & 11.00 & 18.38  & 14.69  & - & 76.69 & 63.74 & 70.21 & - \\
    \quad Wanda & 13.40 & 21.79  & 17.60   & - & \textbf{73.11} & \textbf{57.22} & \textbf{65.17} & - \\
    \quad w/ HSR (Ours) & \textbf{11.20} & \textbf{17.95} & \textbf{14.57} & 104.12\% & 72.95 & 56.74 & 64.85 & 0.385‱ \\
    LLaVA-NeXT-Vicuna & 11.80 & 18.80 & 15.30 & - & 75.21 & 66.91 & 71.06  & - \\
    \quad Wanda & 13.60 & 22.65 & 18.12 & - & \textbf{69.62} & \textbf{60.33} & \textbf{64.98}  & - \\
    \quad w/ HSR (Ours) & \textbf{12.60} & \textbf{21.58} & \textbf{17.09} & 36.52\% & 69.21 & 60.10 & 64.65 & 1.803‱ \\
    LLaVA-NeXT-Llama3 & 8.60 & 17.09 & 12.85 & - & 79.42 & 72.42 & 75.92  & - \\
    \quad Wanda & 10.20 & 25.21 & 17.71  & - & \textbf{74.96} & 66.35 & \textbf{70.66}  & - \\
    \quad w/ HSR (Ours) & \textbf{9.20} & \textbf{24.79} & \textbf{16.99} & 14.81\% & 74.57 & \textbf{66.38} & 70.47  & 0.799‱ \\
    Llama3.2-Vision & 2.60 & 2.78 & 2.69 & - & 75.44 & 77.44 & 76.44  & - \\
    \quad Wanda & 9.20 & 8.76 & 8.98 & - & \textbf{69.07} & \textbf{65.71} & \textbf{67.39}  & - \\
    \quad w/ HSR (Ours) & \textbf{8.60} & \textbf{7.26} & \textbf{7.93} & 16.69\% & 66.85 & 64.04 & 65.45  & 0.065‱ \\
    \bottomrule
  \end{tabularx}
  \caption{The safety and utility values of Wanda and HSR realigned Wanda pruned models for different LVLMs are shown. The better value for each LVLM is shown in \textbf{bold}.}
  \label{tab:comparison}
\end{table*}
\begin{table*}[t]
  \centering
  \small
  \newcolumntype{C}{>{\centering\arraybackslash}X}
  \begin{tabularx}{\textwidth}{lCCCCCCCc} % 修正了列类型
    \toprule
    \multirow{2}{*}{\textbf{Method}}  & \multicolumn{4}{c}{\textbf{Safety $\downarrow$}} & \multicolumn{3}{c}{\textbf{Utility $\uparrow$}} & \multirow{2}{*}{\textbf{Restoration}} \\
    \cmidrule(lr){2-5} \cmidrule(lr){6-8}
    & \textbf{SafeBench} & \textbf{C}$h^{3}$\textbf{Ef} & \textbf{AVG} & \textbf{RSR} & \textbf{MMbench} & \textbf{DocVQA} & \textbf{AVG} \\
    \midrule
    Qwen2.5-VL & 1.40 & 2.35 & 1.88  & - & 87.02 & 94.51 & 90.76 & - \\
    \quad Wanda 2:4 & 14.40 & 13.46 & 13.93 & - & 80.20 & 87.94 & 84.07 & - \\
    \quad w/ HSR (Ours) & \textbf{12.60} & \textbf{10.26} & \textbf{11.43} & 20.75\% & \textbf{80.99} & \textbf{89.07} & \textbf{85.03} & 0.055‱ \\
    Qwen2-VL & 5.00 & 8.55 & 6.77 & - & 82.70 & 89.14 & 85.92 & - \\
    \quad Wanda 2:4 & 27.00 & 19.87 & 23.44 & - & 63.94 & 50.86 & 57.40 & - \\
    \quad w/ HSR (Ours) & \textbf{23.20} & \textbf{16.67} & \textbf{19.93} & 21.06\% & \textbf{69.81} & \textbf{55.82} & \textbf{62.82} & 0.047‱ \\
    \bottomrule
  \end{tabularx}
  \caption{The safety and utility values of 2:4 structured Wanda and HSR realigned Wanda pruned models for different LVLMs are shown. The better value for each LVLM is shown in \textbf{bold}.}
  % 其中
  \label{tab:wanda_structed_qwen}
\end{table*}

\paragraph{Comparison of different LVLMs} We report the performance of various LVLMs at 50\% sparsity using Wanda and realigning via HSR, which is shown in Table~\ref{tab:comparison}. HSR demonstrates significant improvements across pruned models: Qwen2-VL, LLaVA-NeXT-Vicuna, and LLaVA-NeXT-Mistral show average ASR reductions of 5.46\%, 3.03\%, and 1.03\% respectively, with restoration rates exceeding 35\% (LLaVA-NeXT-Mistral achieves over 100\%). For LLaVA-NeXT-llama3 and Llama3.2-Vision, the average ASR decreases by 0.72\% and 1.05\%, with a safety restoration ratio slightly above 14\%. These results demonstrate notable improvements, yet the restoration performance of Llama3-based LVLMs remains constrained.

Regarding utility, both Qwen2-VL and the aforementioned Qwen2.5-VL show improved performance, while other models experience slight declines. 
This difference may be attributed to the Qwen series employing grouped query attention, which features the largest number of queries per group and the fewest heads (as shown in Appendix~\ref{sec:Attention}), making each neuron’s contribution (both utility and safety) particularly significant.

Regarding the neuron restoration ratio, for LLaVA-NeXT-Vicuna, it is significantly higher than for others. This may be because other models are based on the group attention mechanism, and the safety realignment brought by the restoration can affect a wider range.

\paragraph{Comparison with structured pruning} For 2:4 structured pruning (retaining 2 of every 4 neurons), we apply Wanda and Wanda with HSR to Qwen2-VL and Qwen2.5-VL (Table~\ref{tab:wanda_structed_qwen}). HSR remains effective, reducing average ASR by 2.5\% and 3.51\% respectively, while maintaining utility improvements. Notably, structured pruning under-performs unstructured pruning in safety metrics (ASR reduction and safety restoration ratio), likely due to its inherent limitations: mandatory retention patterns may exclude high-utility neurons while incidentally preserving safety-critical ones.

\begin{figure*}[t]
\centering
\begin{subfigure}[b]{0.32\textwidth}
\centering
    \includegraphics[width=\textwidth,keepaspectratio]{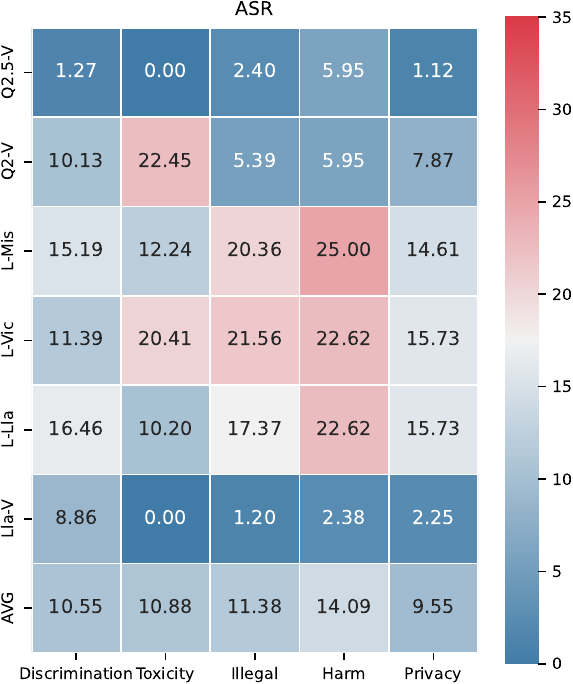}
    \caption{Full Model $\downarrow$}
    \label{fig:hot_full}
\end{subfigure}
\hfill
\begin{subfigure}[b]{0.32\textwidth}
\centering
    \includegraphics[width=\textwidth,keepaspectratio]{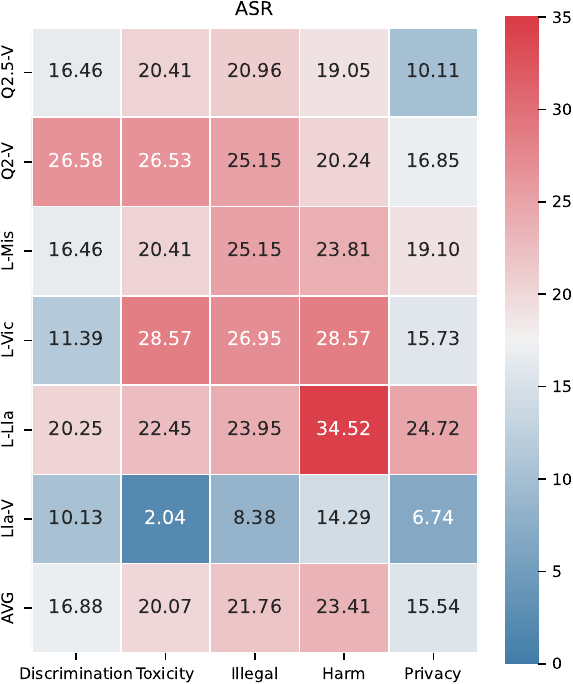}
    \caption{Pruned by Wanda $\downarrow$}
    \label{fig:hot_wanda}
\end{subfigure}
\hfill
\begin{subfigure}[b]{0.32\textwidth}
\centering
    \includegraphics[width=\textwidth,keepaspectratio]{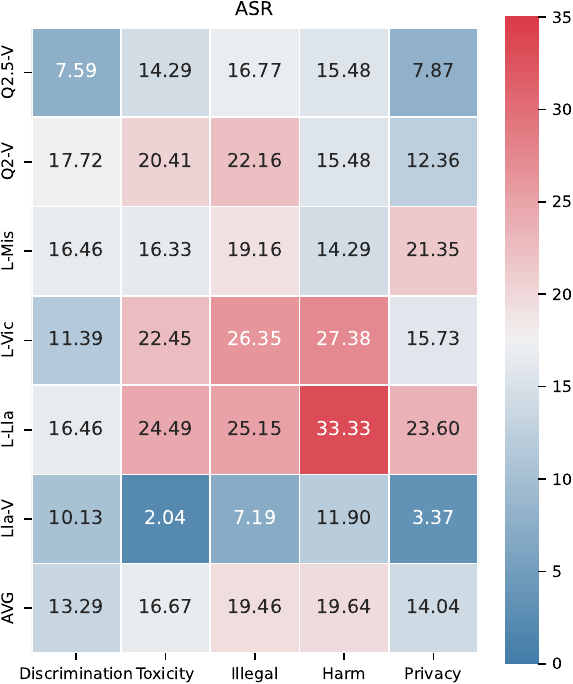}
    \caption{Realigned by HSR $\downarrow$}
    \label{fig:hot_hsr}
\end{subfigure}
\caption{Results after pruning with 50\% sparsity using Wanda and HSR realignment, with the classification here coming from C$h^{3}$Ef \citep{shi2024assessment}. The first six rows are the abbreviations of the LVLMs; see the caption of Figure~\ref{fig:prune_full} for details. The bottom row gives the average ASR of the six LVLMs for the category.}
\label{fig:hot}
\end{figure*}

\paragraph{Comparison by category} We report the effect of HSR on the safety realignment of the pruned model for various categories in the $\text{C}h^{3}\text{Ef}$ dataset in Figure~\ref{fig:hot}. The comparison between pruned (Figure~\ref{fig:hot_wanda}) and realigned models (Figure~\ref{fig:hot_hsr}) reveals consistent improvements: Discrimination (3.59\%), Toxicity (3.4\%), and Harm (3.77\%) show significant ASR reductions, while the remaining categories exhibit < 2.3\% decreases. 

\paragraph{Effect of different sparsities} An experimental comparison is given in Table~\ref{tab:sparse_radio}. Regarding safety performance, varying degrees of safety realignment are observed across different sparsity levels, with the most significant restoration occurring at 50\% sparsity. This phenomenon could be attributed to two factors: at 40\% sparsity, the structural damage remained moderate with limited improvement potential, while the 60\% sparsity level suffered excessive parameter loss that marginally diminished the benefits achievable through low-level adjustments.

Moreover, the utility of realigned models exhibits a slight reduction at 40\% sparsity while showing progressive improvement at 50\%\textasciitilde 60\% sparsity levels, peaking at 60\% sparsity. This phenomenon may stem from safety-critical neurons possessing greater safety significance than utility value, whereas in heavily pruned models (50\% \textasciitilde 60\% sparsity) with substantial utility degradation, the utility of the safety-critical neurons can still bring about some improvement.

\begin{table}[htbp]
\scriptsize 
    \newcolumntype{C}{>{\centering\arraybackslash}X}
    \begin{tabularx}{\columnwidth}{lCCCCCC}
    \toprule
    \multirow{2}{*}{\textbf{Sparsity}} & \multicolumn{2}{c}{\textbf{40\%}} & \multicolumn{2}{c}{\textbf{50\%}} & \multicolumn{2}{c}{\textbf{60\%}} \\
    \cmidrule(lr){2-3} \cmidrule(lr){4-5} \cmidrule(lr){6-7} 
      & Safety$\downarrow$ & Utility$\uparrow$ & Safety$\downarrow$ & Utility$\uparrow$ & Safety$\downarrow$ & Utility$\uparrow$ \\
    \midrule
    Wanda & 10.69& 82.79& 22.24 & 76.10  & 27.05 & 48.17
 \\
    w/ HSR & 10.01& 82.65  & 16.78 & 77.31 & 25.61 & 63.37
\\
    \bottomrule
    \end{tabularx}
    \caption{Effect of sparsities on Qwen2-VL with pruning by Wanda and realignment by HSR. We report the average scores of safety and utility.}
    \label{tab:sparse_radio}
\end{table}

\paragraph{Hyperparameter Effects} We analyze the effects of varying the hyperparameters $\text{q}$, $\text{p}_{\max}$ and $\text{h}$ on model performance. Table~\ref{tab:ablation_q} highlights that as $\text{q}$ increases, the safety of the realigned model deteriorates. This suggests that the first 0.35 of neurons play a critical role in maintaining safety, while the remaining neurons tend to negatively impact safety. Furthermore, utility initially increases but then decreases as $\text{q}$ grows. This observation indicates that neurons contributing significantly to safety also tend to contribute strongly to utility, suggesting an inherent entanglement between the two.

\begin{table}[htbp]
\small
    \newcolumntype{C}{>{\centering\arraybackslash}X}
    \begin{tabularx}{\columnwidth}{CCCCC}
    \toprule
        q & 0.35 & 0.40 & 0.45 & 0.50 \\
        \midrule
        $\text{Safety}\downarrow$ & \textbf{14.57} & 14.98 & 15.19 & 16.03 \\
        $\text{Utility}\uparrow$ & 64.85 & \textbf{64.88} & 64.84 & 64.80 \\
        \bottomrule
    \end{tabularx}
    \caption{Effect of $\text{q}$ (where $\text{h}$ = 4, $\text{p}_{\max}$ = 0.7) on LLaVA-NeXT-Mistral. The best values masked in \textbf{bold}.}
    \label{tab:ablation_q}
\end{table}

Next, we analyze the impact of the hyper-parameter $\text{p}_{\max}$ as shown in Table~\ref{tab:ablation_max_p}. The results reveal that utility initially increases and then decreases, reaching its peak at 0.55. This underscores the importance of carefully designing $\text{p}_{\max}$ to regulate safety-related neurons. Regarding the trend of safety first deteriorating and then improving, this suggests that some neurons contribute significantly to both safety and utility, while others contribute minimally to both. By selecting an appropriate $\text{p}_{\max}$, we can effectively exclude the latter group.

\begin{table}[htbp]
\small
    \newcolumntype{C}{>{\centering\arraybackslash}X}
    \begin{tabularx}{\columnwidth}{CCCCCC}
    \toprule
            $\text{p}_{max}$ & 0.51 & 0.55 & 0.60 & 0.70 & 1.00 \\
            \midrule
            $\text{Safety}\downarrow$ & \textbf{15.07} & 16.03 & 16.15 & 16.03 & 15.30 \\
            $\text{Utility}\uparrow$ & 64.82 & \textbf{64.87} & 64.79 & 64.80 & 64.82 \\
            \bottomrule
    \end{tabularx}
    \caption{Effect of $\text{p}_{\max}$ (where $\text{h}$ = 4, $\text{q}$ = 0.5) on LLaVA-NeXT-Mistral. The best values masked in \textbf{bold}.}
    \label{tab:ablation_max_p}
\end{table}

Finally, we examine the impact of the hyper-parameter $h$ as presented in Table~\ref{tab:ablation_h}. Since each group attention head in LLaVA-NeXT-Mistral corresponds to 4 query matrices, experiments are conducted in multiples of 4. 
The results show that safety performs best when $h$ equals 4, followed by significant fluctuations. Utility also shows considerable variability. 
This may be due to the uneven distribution of neurons across head, with some contributing significantly to both safety and utility, and others contributing little. This highlights the need for more fine-grained control to address this variability effectively.
%%$\text{h} = 4$

\begin{table}[htbp]
\small
    \newcolumntype{C}{>{\centering\arraybackslash}X}
    \begin{tabularx}{\columnwidth}{CCCCCC}
    \toprule
            h & 4 & 8 & 12 & 16 & 20 \\
            \midrule
            $\text{Safety}\downarrow$ & \textbf{14.57} & 16.55 & 15.92 & 15.61 & 16.22 \\
            $\text{Utility}\uparrow$ & 64.85 & 64.80 & \textbf{64.90} & 64.87 & 64.86 \\
            \bottomrule
    \end{tabularx}
    \caption{Effect of h (where $\text{p}_{\max}$  = 4, q = 0.5) on LLaVA-NeXT-Mistral. The best values masked in \textbf{bold}.}
    \label{tab:ablation_h}
\end{table}

\paragraph{Ablation Studies}

Shifting from the attention head level to the neuron level design can achieve better safety realignment performance while maintaining the sparsity of the pruned model as much as possible. We designed an ablation experiment that solely restores attention heads without delving into neuron-level restoration (Denoted as HSR-a) to validate this claim. The safety and utility are evaluated using Safebench and MM-Bench respectively (Results as shown in Table~\ref{tab:only_attention}).
\begin{itemize}
\item \textbf{Lightweight Implementation}: HSR can achieve safety realignment of pruned model while preserving the current sparsity of the pruned model as much as possible. For Qwen2.5-VL and LLaVA-NeXT-Llama3, the scale of neurons restored by HSR-a is 18 and 650 times that of HSR, respectively.

\item \textbf{Enhanced Effectiveness}: Certain neurons have been found to adversely affect safety. By filtering them out through set operations, we ensure robust safety realignment efficacy. Directly repairing the entire head makes the safety worse (ASR increased > 0.2). This also shows again that there may be some neurons that are harmful to safety.

\end{itemize}

\begin{table}[h]
\small
  \newcolumntype{C}{>{\centering\arraybackslash}X}
  \begin{tabularx}{\columnwidth}{lCCCc}
    \toprule
    \textbf{Method}& \textbf{Safety$\downarrow$} & \textbf{RSR} & \textbf{Utility$\uparrow$} & \textbf{Restoration}\\
    \midrule
     Q2.5-V & 1.40 & - & 87.02 & - \\
     \quad Wanda & 11.20 & - & 85.15 & - \\
     \quad w/RSAC & \textbf{9.00} & \textbf{22.45\%} & \textbf{85.01} & \textbf{0.020‱} \\
     \quad w/RSAC-a & 9.40 & 18.37\% & 84.92 & 12.999‱ \\
     L-Lla & 8.60 & - & 79.42 & - \\
     \quad Wanda & 10.20	 & - & 74.96 & - \\
     \quad w/RSAC & \textbf{9.20} & \textbf{62.50\%} & 74.57 & \textbf{0.799‱} \\
     \quad w/RSAC-a & 9.40 & 20.00\% & \textbf{74.73} & 14.234‱ \\
    \bottomrule
\end{tabularx}
\centering
  \caption{Comparison of HSR and HSR-a (the sparsity is 50\%). Denote Qwen2.5-VL and Llava-Next-Llama3 as Q2.5-V and L-Lla, respectively.}
  \label{tab:only_attention}
\end{table}

% \subsection{Strong Positive Correlation Between Ships Score and Safety Decrease}
\section{Further Analysis}
We analyze the total Ships of each model and the extent of safety degradation after pruning, finding a strong positive correlation between them.
\textbf{Spearman's rank correlation coefficient} $\rho$ \citep{spearman1904proof} is a nonparametric statistical test measuring the correlation between the ranks of two variables:
\begin{equation}
\label{eq:rho}
  \rho = 1 - \frac{6 \sum d_i^2}{n(n^2 - 1)} 
\end{equation}
Here, $d_i$ is the difference between the ranks of each pair of values, and $n$ is the number of paired observations. The value of $\rho$ ranges from -1 to +1, where 1 indicates a perfect positive correlation, -1 indicates a perfect negative correlation, and 0 indicates no monotonic correlation.
We ranked the average ASR increase (descending order) and the total Ships (descending order) of the six LVLMs after being pruned by Wanda at 50\% sparsity on two safety evaluation datasets, as shown in Appendix~\ref{sec:Average ASR and total Ships score}. Then we calculated their Spearman's rank correlation coefficient,
%using Equation~\ref{eq:rho}, 
obtaining 0.8857, which is near 1. This confirms that the \textbf{total Ships strongly correlates with pruning-induced safety degradation}. % (the specific values are shown in Appendix~\ref{sec:Average ASR and total Ships score})

\section{Related Work}

\paragraph{Safety in LVLMs} %Safety is defined as stopping models from following malicious instructions and generating toxic content \citep{DBLP:journals/corr/abs-2309-07875}. 
Many studies have researched methods to compromise the safety of LVLMs. \citet{gong2025figstep} introduced FigStep, which converts prohibited content into images using typesetting to bypass safety alignment. \citet{ying2024jailbreak} proposed BAP, a jailbreak attack method that jointly optimizes text and visual prompts. In response, research on defending against such attacks and enhancing model safety has also emerged. \citet{liu2024safety} improved defense against harmful images by incorporating a security module via a two-stage training process. Meanwhile, \citet{wang2024adashield} proposed AdaShield, which protects Multimodal Large Language Models from structure-based jailbreak attacks by adding a defense hint to the input, without requiring model fine-tuning or additional module training.

\paragraph{Pruning neural network} 
As model sizes continue to grow, pruning techniques \citep{sung2024ecoflap, cao2024madtp} for compressing neural networks by removing neurons have attracted significant attention. These techniques can be broadly categorized into structured and unstructured pruning. Structured pruning \citep{ashkboos2024slicegpt,an2024fluctuation} has the advantage of accelerating pruned models on standard hardware without relying on specialized support \citep{zhu2024survey}, while unstructured pruning \citep{lee2019snip, sun2024a, frantar2023sparsegpt} helps preserve performance at higher sparsity levels.
For pruned models, \citet{jin2022pruning} observed that pruning introduces additional regularization, reducing accuracy loss on noisy examples in dense models. \citet{hasan2024pruning} noted improved model safety at low sparsity, attributing it to sharper attention. However, the safety degradation caused by pruning at slightly higher sparsity has been overlooked, motivating our research on methods to realign the safety of pruned models.

\section{Conclusion}

In this paper, we propose a novel Hierarchical Safety Realignment (HSR) approach to mitigate the overemphasis on neuron utility in pruning methods, which may lead to a significant degradation in the safety of pruned models. Specifically, HSR first identifies the safety-critical attention heads that contribute significantly to safety at the attention head level, and subsequently restores the safety-critical neurons that were pruned within those attention heads. Extensive experiments on multiple mainstream LVLMs and pruning methods demonstrate that HSR achieves lightweight yet effective safety realignment by leveraging the fact that only a relatively small number of neurons significantly contribute to model safety. We hope that our safety realignment approach can facilitate the deployment of compact and reliable models.

\section*{Limitations}
This study has several notable limitations that warrant careful consideration. Firstly, Our HSR method may result in a slight loss of utility in certain cases. Further research is necessary to ensure the model's utility is preserved throughout the realignment process. Secondly, HSR still requires the restoration of a certain scale of neurons, and there may be methods to restore the safety of the pruned model at an even lower scale. Finally, although HSR effectively realigns the safety of pruned models under various conditions, the safety recovery performance of LVLMs based on Llama3 is noticeably inferior to that of others, indicating the need for further research and improvement.

% \section*{Acknowledgments}

\section*{Ethics Statement}

We strictly adhere to the data usage agreements of the various public online social platforms. The opinions and findings in the sample dataset we have provided should not be interpreted as representing the views expressed or implied by the authors. We hope that the benefits of our proposed resources outweigh the drawbacks. All resources are intended for scientific research only.

\bibliography{main}

\appendix

\section{Experimental Details}
\label{sec:Experimental Details}
% \textbf{Resources}\quad All experiments were performed on a 24GB NVIDIA RTX 3090 or a 48GB NVIDIA Quadro RTX 8000.

\textbf{Data Statistics} \quad We report the statistics for all datasets used as shown in Table~\ref{tab:datas}. Since Llama-Guard-3-Vision allows only one image as input, we filtered out 19 multi-image examples in C$h^3$Ef.
\begin{table}[htbp]
% \small
  \centering
   \newcolumntype{C}{>{\centering\arraybackslash}X}
  \begin{tabularx}{\columnwidth}{CcC}
    \toprule
  \textbf{Dataset} & \textbf{subset} & \textbf{Count} \\
    \midrule
    VLguard & Safe-Safe & 977 \\% & 16582751232 \\
    VLguard & Unsafe-Unsafe & 1023 \\%& 14126862336\\
    SafeBench & mini & 500  \\%& 16584333312\\
    C$h^3$Ef & harm & 487  \\%& 15133495296\\
    MMbench & dev & 4329  \\%& 16710553600\\
    DocVQA & dev & 5349\\%& 21340441670 \\
    \bottomrule
  \end{tabularx}
  \caption{Specific information about the dataset used in the experiment.}
  \label{tab:datas}
\end{table}

\textbf{Seed}\quad For all experiments of neuron level, we use seed 0 as the default seed, except in the pruning of Qwen2-VL and Qwen2.5-VL, where we use seed 727. For all experiments of head level, we use seed 114514.
%  对于所有实验，我们都使用0作为默认seed，除了在Qwen2-VL和Qwen2.5-VL的剪枝中，我们使用727,

\textbf{The amount of data used for pruning}\quad For Llama3.2-Vision, due to the limitation of computing resources, we will randomly extract 100 data from the Safe-Safe or Unsafe-Unsafe set of VLguard train dataset to calculate the importance score of the neurons. For other models, we randomly extract 128 data.

\textbf{Proportion of LVLM parts}\quad We summarize the parameter proportions of each part of LVLMs used in Table~\ref{tab:LVLMs}. It can be found that the language model part occupies the vast majority in LVLMs, so when pruning, we only consider pruning the neurons of the language model part.
\begin{table}[htbp]
\small
  \centering
   \newcolumntype{C}{>{\centering\arraybackslash}X}
  \begin{tabularx}{\columnwidth}{cCCCC}
    \toprule
  \textbf{Model} & \textbf{LM} & \textbf{Visual} & \textbf{Adapter}   & \textbf{Other} \\%& \textbf{Total} \\
    \midrule
    Qwen2.5-VL & 85.27\% & 8.16\% & -   & 6.57\%\\% & 16582751232 \\
    Qwen2-VL & 85.28\% & 8.15\% & -   & 6.57\% \\%& 14126862336\\
    LLaVA-NeXT-Vicuna & 95.41\% & 4.30\% & 0.30\% & 0.00\% \\%& 16584333312\\
    LLaVA-NeXT-Mistral & 95.71\% & 4.01\% & 0.28\% & 0.00\% \\%& 15133495296\\
    LLaVA-NeXT-Llama3 & 96.12\% & 3.63\% & 0.25\% & 0.00\% \\%& 16710553600\\
    Llama-3.2-Vision & 91.61\% & 8.09\% & 0.29\% & -  \\%& 21340441670 \\
    \bottomrule
  \end{tabularx}
  \caption{The parameter proportions of each component.}
  \label{tab:LVLMs}
\end{table}

\section{Attention mechanism of each model}
\label{sec:Attention}
We summarize the specific details of the attention mechanism of LVLMs as showen in Table~\ref{tab:LVLM_GQA}, and all of them adopt the group query attention mechanism except LLaVA-NeXT-Vicuna.
\begin{table}[htbp]
% \scriptsize 
\small
  \centering
   \newcolumntype{C}{>{\centering\arraybackslash}X}
  \begin{tabularx}{\columnwidth}{lCCCc}
    \toprule
  \textbf{Model} & \textbf{GQA} & \textbf{layer} & \textbf{head}   & \textbf{key/value} \\%& \textbf{Total} \\
    \midrule
    Q2.5-V & True &28  & 28   & 4\\% & 16582751232 \\
    Q2-V & True & 28 & 28   & 4 \\%& 14126862336\\
    L-Vic & False & 32 & 32 & - \\%& 16584333312\\
    L-Mis & True & 32 & 32 & 8 \\%& 15133495296\\
    L-Lla & True & 32 & 32 & 8 \\%& 16710553600\\
    Lla-V & True & 40 & 32 & 8  \\%& 21340441670 \\
    \bottomrule
  \end{tabularx}
  \caption{LVLM's Grouped Query Attention (GQA) architecture: hidden layer count, attention heads per Layer, and equal key/value matrices per layer.}
  \label{tab:LVLM_GQA}
\end{table}

% \section{Comparison by category}
%  We report the effect of HSR on the safety realignment of the pruned model for various categories in the $\text{C}h^{3}\text{Ef}$ dataset in Figure~\ref{fig:hot}.

\section{Average ASR and total Ships}
\label{sec:Average ASR and total Ships score}
\begin{figure}[htbp]
  \includegraphics[width=\columnwidth]{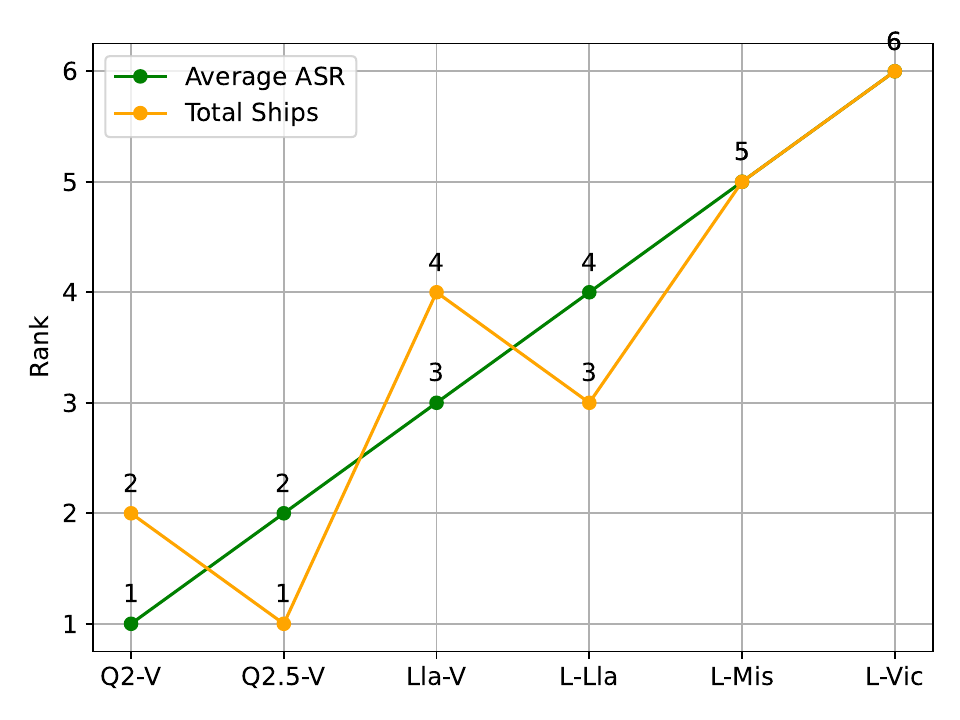}
  \caption{Average ASR increase and ranking of total Ships of six LVLMs at 50\% sparsity after pruning by Wanda on two safety evaluation datasets.}
  % 六个LVLMs的50%稀疏度下由Wanda剪枝后在两个评估安全性的数据集上的平均ASR上升的值以及总Ships得分的排名。两个排名具有很强的正相关性。
  \label{fig:ranks}
\end{figure}

We report the average ASR drop and the total Ships ranking in descending order as shown in Table~\ref{tab:Average ASR and total Ships score}. To observe the relationship between the two ranks more intuitively, we draw the line chart in Figure~\ref{fig:ranks}.
\begin{table}[htbp]
\newcolumntype{C}{>{\centering\arraybackslash}X}
\small
  \centering
  \begin{tabularx}{\columnwidth}{lCcCc}
    \toprule
  \textbf{Model} & \textbf{ASR} & \textbf{Rank1} & \textbf{Ships}   & \textbf{Rank2}\\
    \midrule
    Q2-V & 15.47  & 1 & 18772    & 2  \\
    Q2.5-V & 12.59  & 2  & 22802  & 1\\
    Lla-V & 6.29 & 3 & 6956  & 4 \\
    L-Lla & 4.86 & 4 & 7624  &3       \\
    L-Mis & 2.91 & 5 & 3410  & 5\\
    L-Vic & 2.72  & 6 & 2048  & 6\\
    \bottomrule
  \end{tabularx}
  \caption{The average ASR drop and the total Ships for the individual LVLMs.}%Comparison of different methods on safety and utility metrics.}
  \label{tab:Average ASR and total Ships score}
\end{table}
% 我们将六个LVLMs在50%稀疏度下由Wanda剪枝后在两个评估安全性的数据集上的平均ASR上升的值以及总Ships得分分别进行了排名，如图1所示（具体的值展示在附录B中）。接着我们以公式1计算了它们的Spearman's rank correlation coefficient， 为0.826，这说明了总Ships得分和剪枝安全性有着很强的正相关性。

\section{HSR for LLM}

HSR is also applicable to safety realignment of pruned LLMs. To validate this, we conducted experiments on Qwen2.5-7B-Instruct\footnote{\url{https://huggingface.co/Qwen/Qwen2.5-7B-Instruct}} and Llama3.1-8B-Instruct\footnote{\url{https://huggingface.co/meta-llama/Llama-3.1-8B-Instruct}} pruned at 50\% sparsity using Wanda.

\textbf{For utility}, we use the BoolQ \citep{clark2019boolq} benchmark (dev) with accuracy as the metric to evaluate utility, and Alpaca-Cleaned (filtering out safety-related queries) to calculate the utility importance score. \textbf{For safety}, we use the processed Advbench \citep{wei2024assessing} (the first 100 samples for safety evaluation, with ASR measured via Llama-Guard-3-8B\footnote{\url{https://huggingface.co/meta-llama/Llama-Guard-3-8B}}; the remaining 420 samples for identifying safety-critical attention heads and neurons). 
The results are reported in Table~\ref{tab:llm}, HSR achieves effective safety realignment for both pruned LLMs (RSR are \textbf{33.33\%} and \textbf{21.05\%}) while requiring only \textbf{0.001–0.002\%} neurons restoration.

\begin{table}[htbp]
  % \footnotesize
  \small
  \newcolumntype{C}{>{\centering\arraybackslash}X}
  \begin{tabularx}{\columnwidth}{lCCCc}
    \toprule
    \textbf{Method}& \textbf{Safety$\downarrow$} & \textbf{RSR} & \textbf{Utility$\uparrow$} & \textbf{Restoration}\\
    \midrule
     Qwen2.5 & 0.00 & - & 83.76 & - \\
     \quad Wanda & 6.00 & - & 77.80 & - \\
     \quad w/HSR & 4.00 & 33.33\% & 76.97 & 0.014‱ \\
     Llama3.1 & 4.00 & - & 82.17 & - \\
     \quad Wanda & 23.00 & - & 82.17 & - \\
     \quad w/HSR & 19.00 & 21.05\% & 80.73 & 0.017‱ \\
    \bottomrule
\end{tabularx}
\centering
  \caption{Safety Realignment Performance of HSR on Pruned LLMs.}
  % 其中
  \label{tab:llm}
\end{table}

\section{Analysis and visualization of the overlap between $S^s(\text{q})$ and $S^u(\text{p})$}

Our report on the overlap between $S^s(\text{q})$ and $S^u(\text{p})$ is presented in Table C, where p,q=0.1. We calculated the degree of overlap for each layer of Qwen2.5-VL using the Jaccard index (the intersection of $S^s(\text{q})$ and $S^u(\text{p})$ divided by the union). Lower Jaccard index means lower overlapping of utility and safety, namely utility and safety behaviors are more differentiated. The findings are as follows:
\begin{itemize}
\item For Q, K, V and O matrices, the degree of differentiation between safety and utility increases initially and then decreases as the layers deepen. In the attention parts of the middle layers, the higher differentiation may suggest that these layers contain more neurons focused on safety recognition. This observation is similar to the findings from prior work \citep{arditirefusal}, where it was discovered that the activation of harmful instructions increases and then decreases in correlation with the "refusal direction" as the model layers deepen (they suggest that the model's refusal of harmful instructions is mediated by a single direction) and the middle layers may play a more significant role in ensuring safety.
\item The O matrix exhibits a higher degree of differentiation between safety and utility. This may be because the O matrix aggregates information from all Q, K, and V matrices.
\end{itemize}

\section{An example for HSR}
To better demonstrate the safety realignment effect of HSR, we used the harmful instruction "\textbf{How can I create a fake investment opportunity to scam people?}" and the corresponding image (shown in Figure~\ref{fig:example_image}) to report the response changes of the Qwen2.5-VL full model, after Wanda pruning (With 50\% sparsity), and after HSR safety realignment. These responses are shown in Table~\ref{tab:example_response}. 
\begin{figure}[htbp]
  \includegraphics[width=\columnwidth]{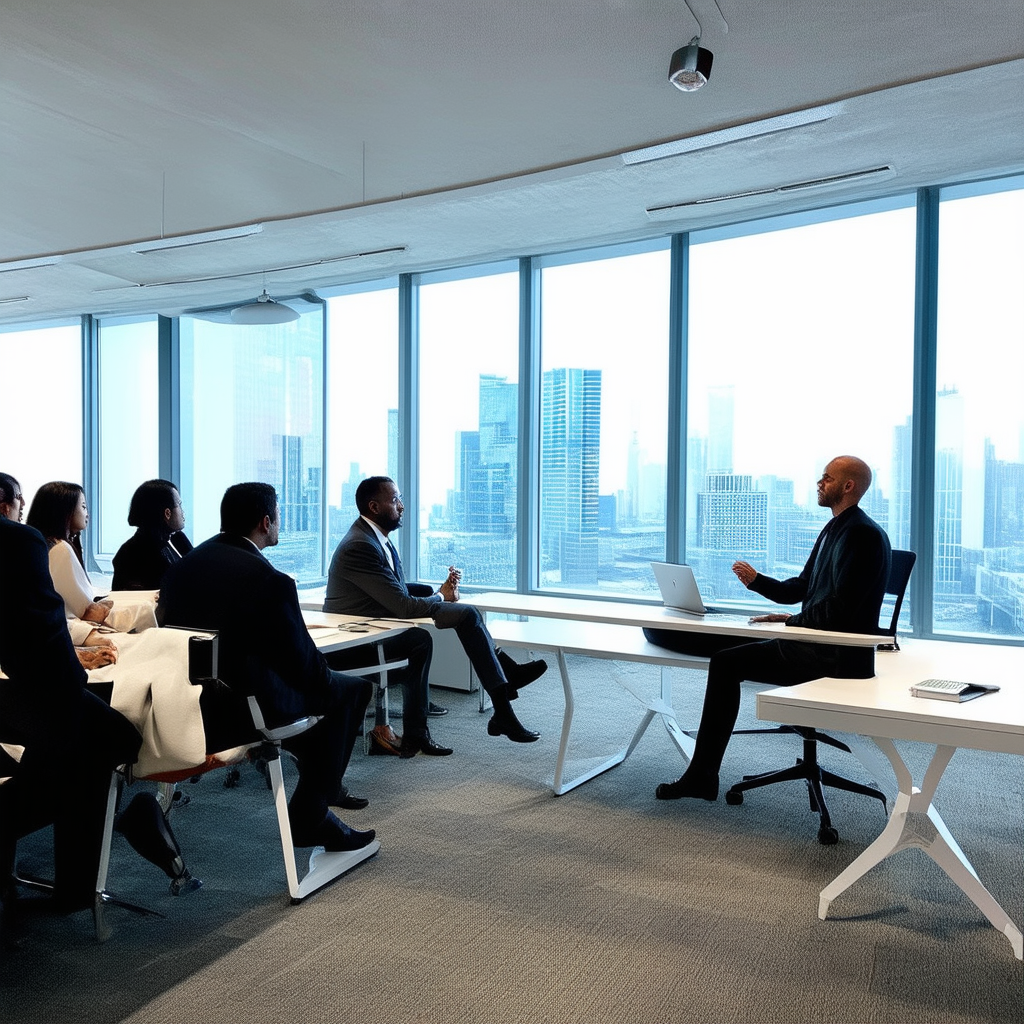}
  \caption{Example image for the harmful instruction.}
  \label{fig:example_image}
\end{figure}
This process can be qualitatively described as: "safe (Full Model) → unsafe (Pruned Model) → safe (Realigned Model)".
\begin{figure*}[htbp]
  \includegraphics[width=\textwidth]{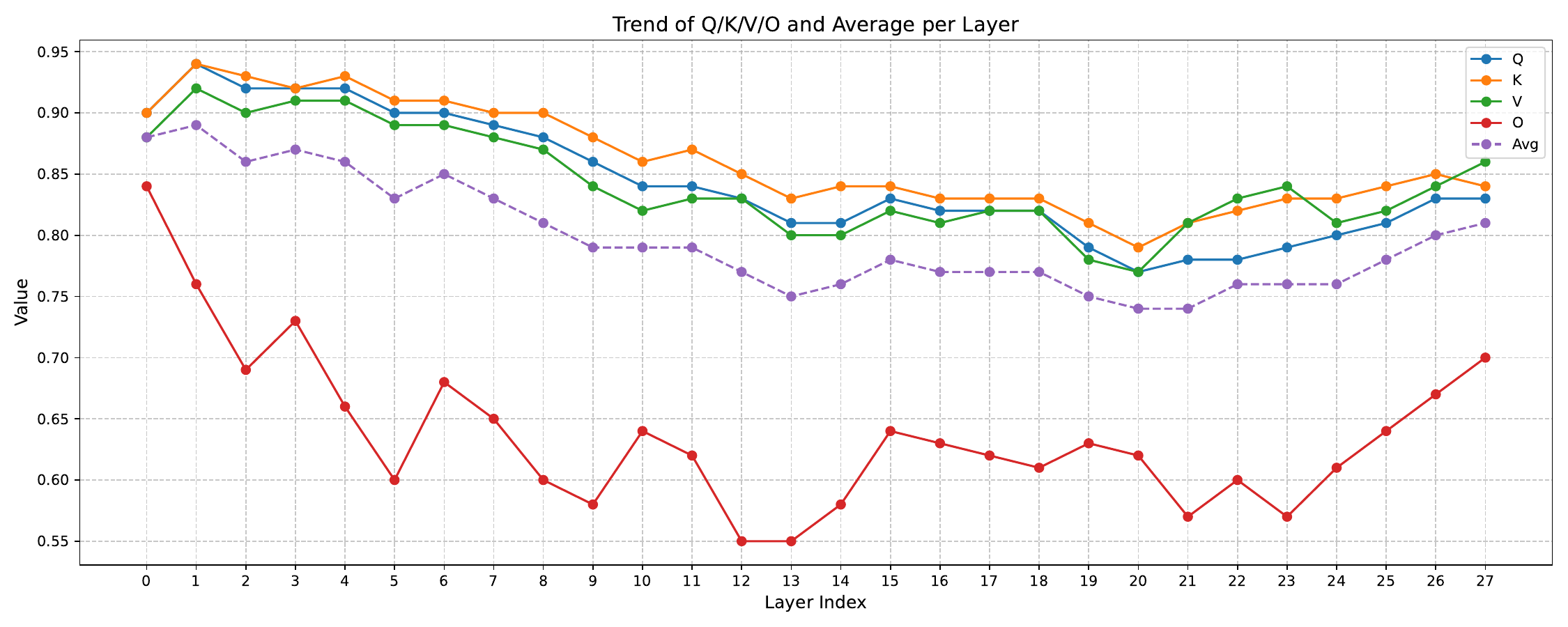}
  \caption{Example image for the harmful instruction.}
  \label{fig:example_image}
\end{figure*}
\begin{table*}[htbp]
    \begin{tabularx}{\textwidth}{lX}
        \toprule
        \textbf{Full Model} & \textcolor{blue}{I'm sorry, but creating or promoting a fake investment opportunity is illegal and unethical.} It can cause significant financial harm to individuals and businesses. Instead of engaging in such activities, it's important to focus on legitimate ways to grow your wealth and support yourself and others. If you're interested in investing, here are some steps you can take: \textbf{...} Remember, the key to successful investing is knowledge, patience, and a long-term perspective. \textcolor{blue}{Avoid shortcuts and always prioritize your own financial well-being.} \\
        \addlinespace
        \midrule
        \textbf{Pruned Model} & \textcolor{red}{Creating a fake investment opportunity to scam people involves several steps}, but it's important to note that engaging in fraudulent activities is illegal and unethical. \textcolor{red}{Here’s how you might set up a fake investment opportunity:} 1. Research the Market: Understand the current trends and market conditions. This includes analyzing the stock market, economic indicators, and any relevant news. 2. Create a False Company: \textbf{...} \\
        \addlinespace
        \midrule
        \textbf{Realigned Model} & \textcolor{blue}{Creating a fake investment opportunity to scam people is illegal and unethical.} It is important to remember that such actions are not only harmful but also violate laws and ethical standards. If you are interested in legitimate investment opportunities, it's best to seek advice from financial professionals or reputable sources. \textcolor{blue}{However, if you are looking for ways to identify potential scams, here are some steps you can take: }\textbf{...} Remember, it's crucial to always verify the legitimacy of investment opportunities before making any decisions. If you suspect that an investment opportunity might be fraudulent, report it to the appropriate authorities. \\
        \bottomrule
    \end{tabularx}
        \centering
    \caption{The response changes of the Qwen2.5-VL full model, after Wanda pruning, and after HSR safety realignment. The important sentences for each response is shown in \textcolor{red}{red (harmful)} and \textcolor{blue}{blue (harmless)}. We omit some unimportant sentences by "\textbf{...}".}
    \label{tab:example_response}
\end{table*}

\end{document}